\documentclass[conference]{IEEEtran}
\IEEEoverridecommandlockouts
\usepackage{cite}
\usepackage{amsmath,amssymb,amsfonts}
\usepackage{algorithmic}
\usepackage{graphicx}
\usepackage{textcomp}
\usepackage{xcolor}
\def\BibTeX{{\rm B\kern-.05em{\sc i\kern-.025em b}\kern-.08em
    T\kern-.1667em\lower.7ex\hbox{E}\kern-.125emX}}
\begin{document}

\title{Non-Compression Auto-Encoder for Detecting Road Surface Abnormality via Vehicle Driving Noise\\
\thanks{This work has been submitted to the IEEE for possible publication. Copyright may be transferred without notice, after which this version may no longer be accessible.}
}

\author{\IEEEauthorblockN{YeongHyeon Park\thanks{Corresponding author: yeonghyeon@sk.com}}
\IEEEauthorblockA{\textit{SK Planet Co., Ltd.}\\
Seongnam, Republic of Korea \\
yeonghyeon@sk.com}
\and
\IEEEauthorblockN{JongHee Jung}
\IEEEauthorblockA{\textit{SK Planet Co., Ltd.}\\
Seongnam, Republic of Korea \\
jonghee.jung@sk.com}
}

\maketitle

\begin{abstract}
Road accidents can be triggered by wet roads because it decreases skid resistance. To prevent road accidents, detecting abnormal road surfaces is highly useful. In this paper, we propose the deep learning-based cost-effective real-time anomaly detection architecture, naming with non-compression auto-encoder (NCAE). The proposed architecture can reflect forward and backward causality of time-series information via convolutional operation. Moreover, the above architecture shows higher anomaly detection performance of published anomaly detection models via experiments. We conclude that NCAE is a cutting-edge model for road surface anomaly detection with 4.20\% higher AUROC and 2.99 times faster decisions than before.
\end{abstract}


\section{Introduction}
\label{sec:introduction}
Comprehending the road surface status is highly helpful for preventing car accidents. In prior research, Hall et al. have shown the wet-weather not only increases water film thickness of the road but also reduces friction coefficient between tire and surface \cite{hall2009guide}. The above means vehicle can be slip on road by lower friction surface also same meaning as lower skid resistance.

In another report, CM McGovern et al. have analysed that the more than 20\% and 35\% road accidents occurred during wet weather than dry weather at the Virginia and New York respectively \cite{mcgovern2011state}. Based on the above, recognizing the dangers during driving on the road can be the best way to prevent and respond to an accident. Thus, we propose a deep learning based anomaly detection method that assists people to more accurately recognize the road surface status.

\section{Related work}
\label{sec:related_work}

Aggregating the time information helps to generate meaningful result for handling time series data. As an effort for example through neural networks, there is a causal reflection model as known as recurrent neural network (RNN) \cite{mikolov2010recurrent}.

The advanced researches for processing time series information, there are two types of neural network are exists. One of them is RNN based model. We refer to recurrent style neural networks collectively as RNN including basic RNN, long short-term memory (LSTM) \cite{graved2012lstm}, and some others. The other one is based on convolutional neural networks (CNN). For example, FARED and HP-GAN are developed based on RNN and CNN respectively for anomaly detection of time series data \cite{park2018fared, park2020hpgan}.

In the prior researches have shown deep learning based state-of-the-art anomaly detection models. The FARED achieved the highest level of area under receiver operating characteristic curve (AUROC) of anomaly detection in surface mounted device assembly machine \cite{park2018fared}. The above model, FARED, is a proper example of reflecting the causality of time series information using RNN. 

However, FARED can reflects causality only forward way because it does not consider the bidirectional method for model construction \cite{schuster1997birnn}. The reason for not considering bidirectional RNN, it highly increases conputational complexity, including depth of layers of the neural network, and time consumption for training and test procedure.

For easing the above limitation, CNN can be considered to construct the neural network architecture as shown in HP-GAN case \cite{park2020hpgan}. The convolutional filter slides on the input data and aggregates the spatial information for generating results. The other case in the above process, when the input data includes time series information, the convolutional filter aggregates and reflects the time information at generating output. Moreover, convolutional filter can reflect forward and backward causality without any settings such as considering bidirectional method.

\section{Proposed approach}
\label{sec:proposed_approach}
We present the CNN-based anomaly detection model in this section. The complex and heavy architecture will consume more time for decisions and it causes information delay. Delayed information does not useful for preventing accidents. Thus, we consider the simple lightweight architecture for real-time processing. 

Prior research, CNN based time series anomaly detection, uses encoder-decoder architecture. Encoder compresses the input data to latent space with lower dimension, and decoder reconstruct them to higher original input dimension. The ability to encoder the input into elaborate low-dimensional vectors and the ability to elaborately restore the encoded information back to its original dimension is likely to be superior as the model size increases. Thus, encoder-decoder style architecture needs more complex neural network architecture, and that makes construct simple architecture difficult.

\begin{figure}[h]
    \begin{center}
		\includegraphics[width=0.60\linewidth]{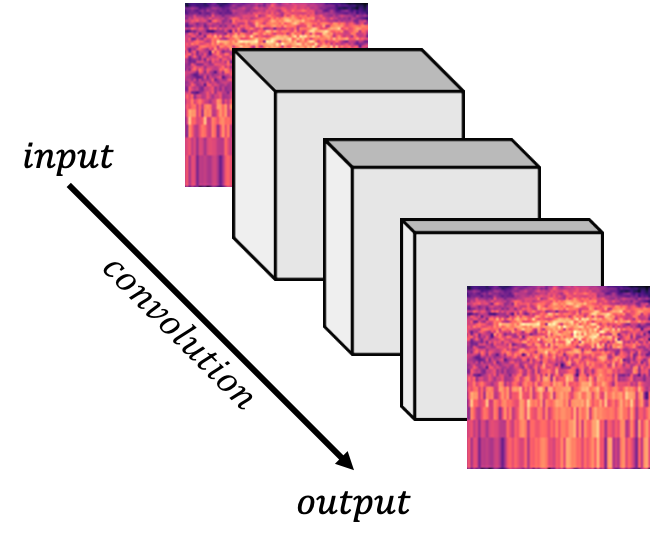}
	\end{center}
	\vspace*{-5mm}
	\caption{Architecture of the non-compressed auto-encoder.}
	\label{fig:ncae}
\end{figure}

Thus, we propose a novel non-compressional style neural network architecture as shown in Figure~\ref{fig:ncae} with convolutional layers. We named the above CNN-based generative neural network as a non-compression auto-encoder (NCAE). The NCAE has only a transition process from input to output, not the encoding and decoding process. According to the above method, there is no need to complicate the neural network architecture in order to perform the encoding and decoding elaborately, and the time consumption for training and test can be shortened.

Each convolutional layer is designed to remember the appropriate information at the filter for converting inputs to outputs. The NCAE leans to replicating the normal data only via combining each filter information. We use this property for anomaly detection. For example, when the abnormal data input to NCAE, abnormalities can be detected by the larger distance between input and output, because NCAE reproduces every input in a normal style.

\section{Experiments}
\label{sec:experiments}
In this section, we present the dataset and experimental results for proving the validity of proposed model NCAE. 

\subsection{Dataset}
\label{sec:dataset}
\begin{figure}[h]
    \begin{center}
        \begin{tabular}{cc}
		    \includegraphics[width=0.45\linewidth]{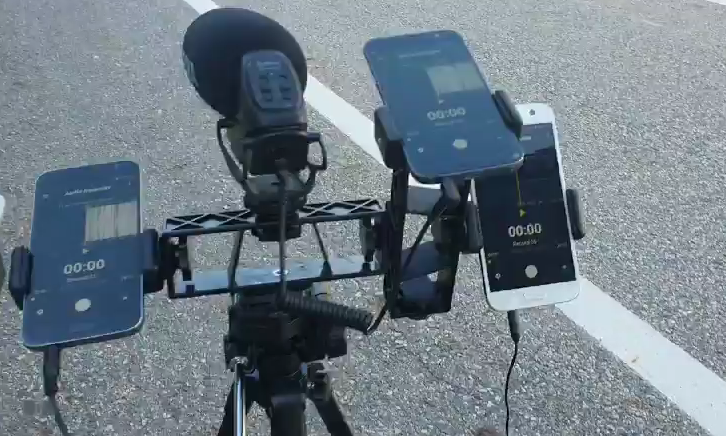} &
		    \includegraphics[width=0.45\linewidth]{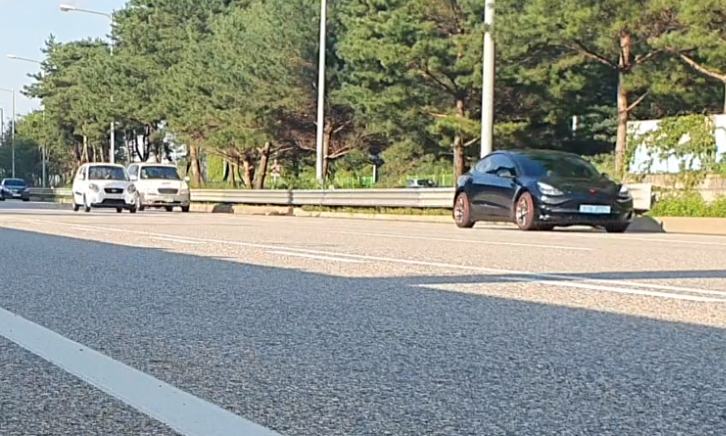}
		\end{tabular}
	\end{center}
	\vspace*{-5mm}
	\caption{Environment for collecting the dataset. Each figure shows the microphone and road sequentially.}
	\label{fig:dataset}
\end{figure}

First of all, we have collect the dataset of vehicle driving noise at Yongin-Seoul expressway with two weather condition, dry and wet. We collect the data to each condition as 20 minute and we split them to training and test set. Figure~\ref{fig:dataset} shows the data collecting environment.

\subsection{Anomaly detection}
\label{sec:anomaly_detection}
We use two models, FARED and NCAE, for experiment in this paper. In order to create a lightweight model, we limit the depth of layers as 3 to construct the above two models. Other CNN-based generative models, including HP-GAN, are excluded from the experiment due to the use of the encoder-decoder structure as described in Section~\ref{sec:related_work}. For comparing the performance, we conduct Monte Carlo estimation to FARED and NCAE with various hyperparameter \cite{kroese2014monte}.

We use preprocessing method, including MFCC feature extraction, already presented in prior research \cite{park2018fared}. We set the noise of dry condition as a normal data, and wet condition as a abnormal data. For detecting abnormality, we conduct training the two generative neural network, FARED and NCAE, with the normal data, and conduct validation process with normal and abnormal data. The loss function for optimization is Euclidean distance between the input $X$ and output $\hat{X}$ of the neural network, as shown in \ref{eq:target}. 

\begin{equation}
    \label{eq:target}
    \mathcal{L} = ||X - \hat{X}||_{2}
\end{equation}

We decide the decision boundary $\theta$ using $\mu$ and $\sigma$ from the training data and loss function referring to Tukey's fences \cite{tukey1977exploratory}, as shown in \ref{eq:decision}.

\begin{equation}
    \label{eq:decision}
    \theta = \mu + (1.5 * \sigma)
\end{equation}

The measured performance is presented in Table~\ref{table:comparison}. AUROC and time consumption are used as indicator for comparing the performance. The higher mean and lower standard deviation represent the higher anomaly detection performance (or higher time efficiency) and higher stability respectively. 

Referring to Table~\ref{table:comparison}, NCAE shows 4.20\% higher anomaly detection performance and 2.99 times faster computation. Also, the standard deviation of NCAE to two performance indicator is lower than FARED, that means NCAE works highly stably.

\begin{table}[h]
    \centering
    \small
    \caption{Measured performance of experiment for comparing between FARED \cite{park2018fared} and NCAE. Each performance is provided with a mean $\pm$ standard deviation form. \\}
    
    \begin{tabular}{ccc}
        \hline
        \textbf{Model} & \textbf{AUROC} & \textbf{Time Consumption} \\ 
        \hline
        FARED & 0.95566 $\pm$ 0.03351 & 0.02590 $\pm$ 0.00105 \\ 
        NCAE & 0.99582 $\pm$ 0.00947 & 0.00867 $\pm$ 0.00012 \\ 
        \hline
    \end{tabular}
    
    \label{table:comparison}
\end{table}

\begin{figure}[ht]
    \begin{center}
        \begin{tabular}{cc}
    		\includegraphics[width=0.45\linewidth]{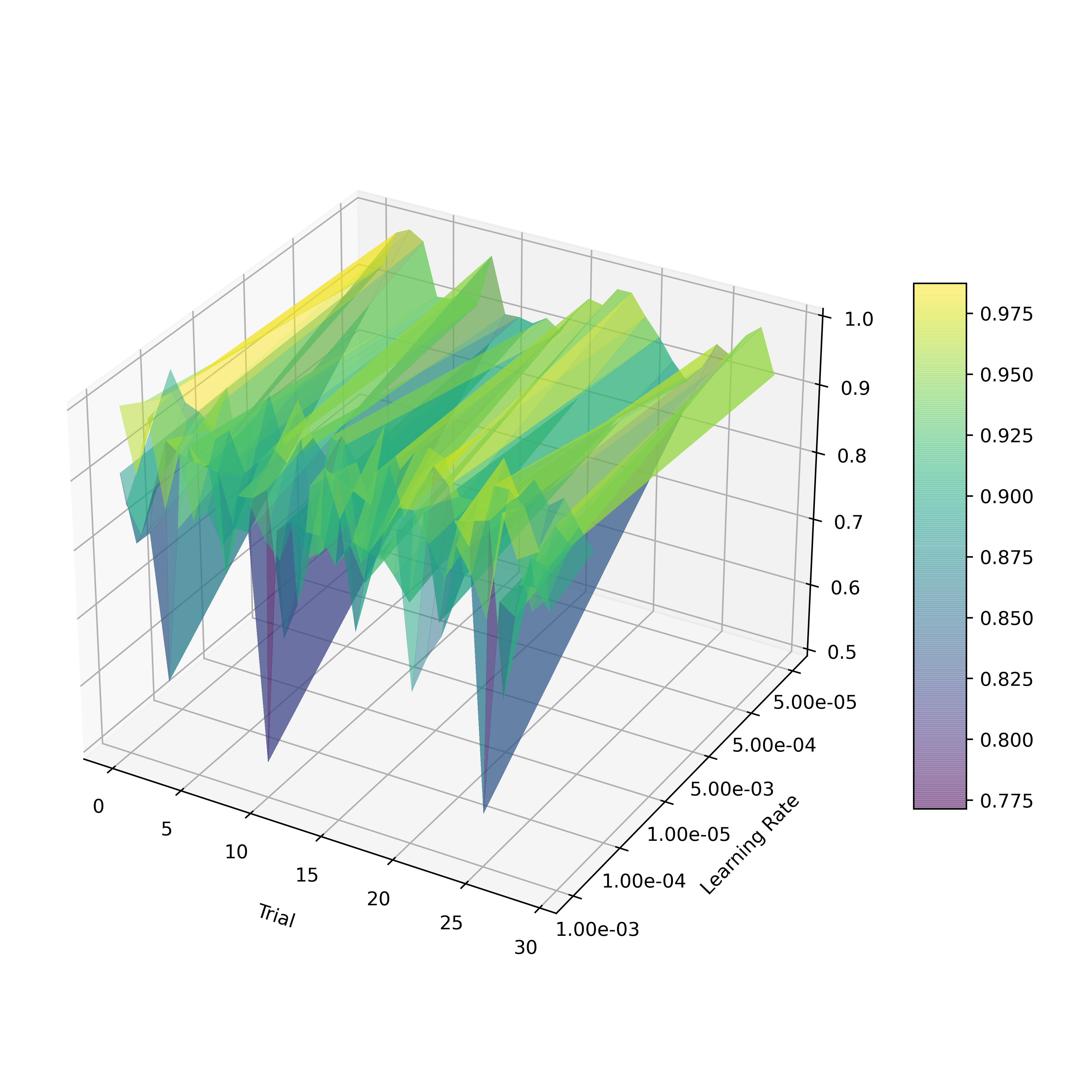} & 
    		\includegraphics[width=0.45\linewidth]{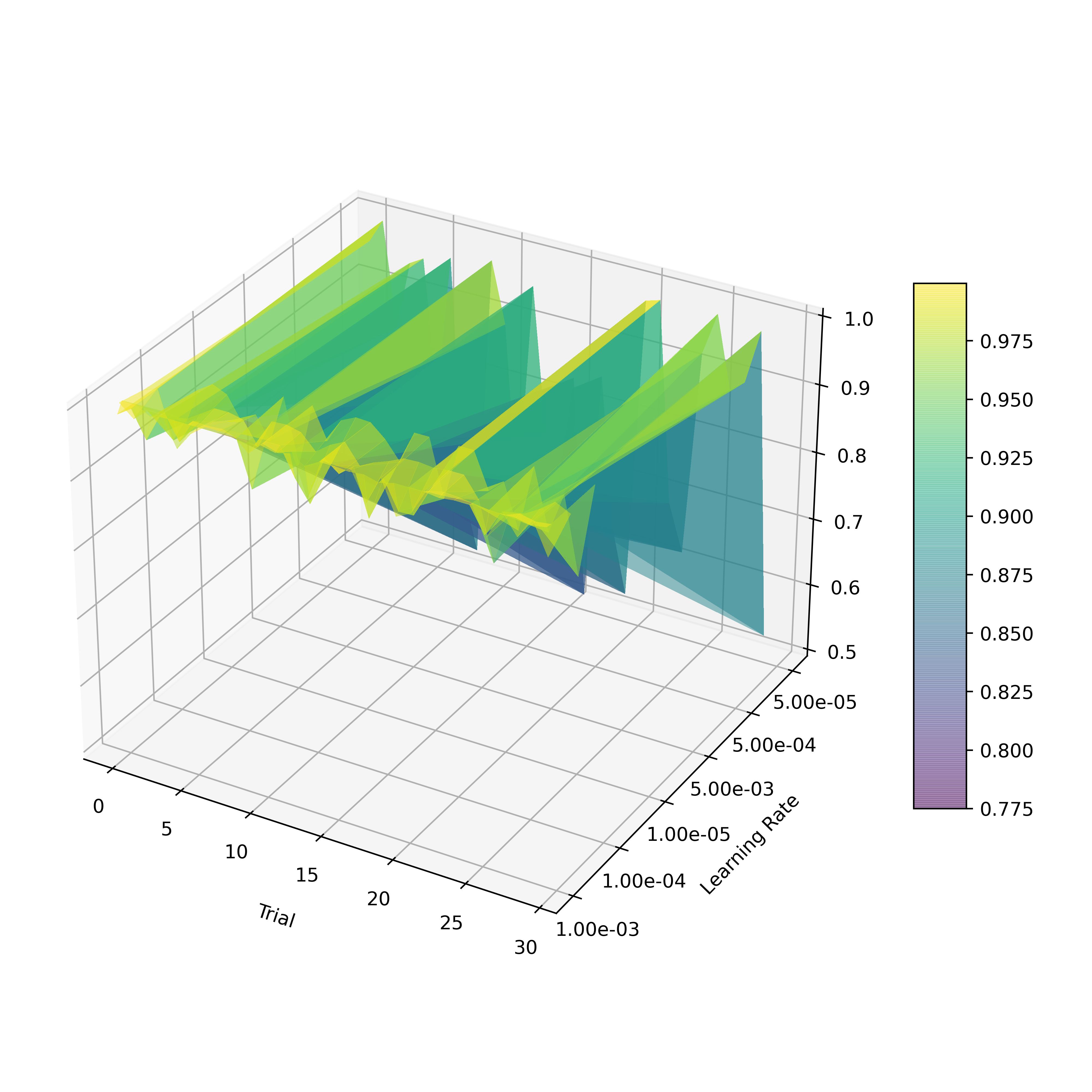} \\
    		FARED & NCAE (kernel size: 3) \\
    		\includegraphics[width=0.45\linewidth]{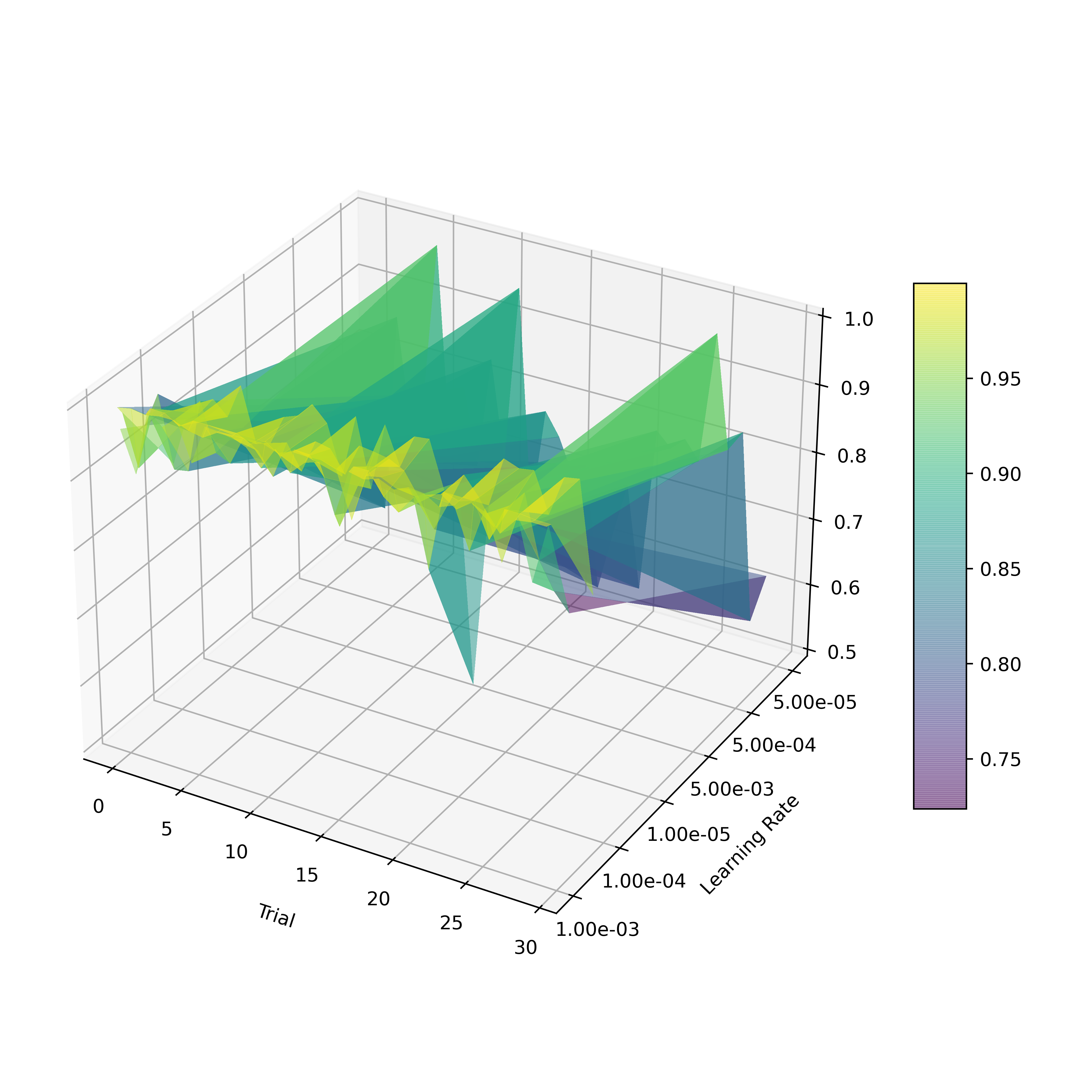} &
    		\includegraphics[width=0.45\linewidth]{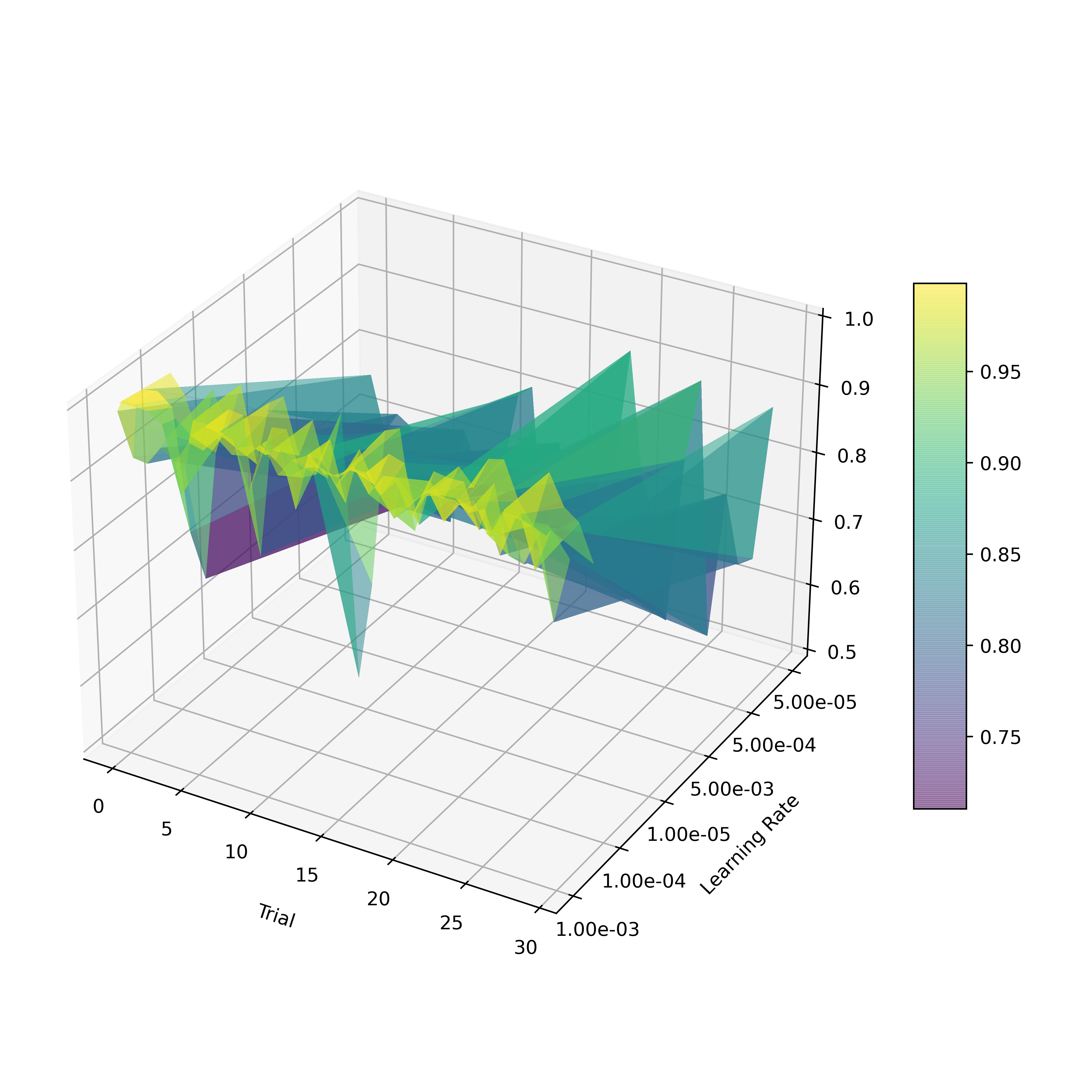} \\
    		NCAE (kernel size: 5) & NCAE (kernel size: 7)
		\end{tabular}
	\end{center}
	\vspace*{-5mm}
	\caption{The first subfigure shows the surface of AUROC for FARED \cite{park2018fared}, and the others show the surface of NCAE. Because of the adjustable hyperparameter of FARED is learning rate only, the only one AUROC surface is generated. However, the diverse surfaces are generated after training NCAE because of the varied kernel size 3, 5, and 7.}
	\label{fig:surfaces}
\end{figure}

We also present the measured AUROC for each hyperparameter as a surface form in Figures~\ref{fig:surfaces}. In figures~\ref{fig:surfaces}, surface of the FARED has fluctuated form for overall space. Refer that, more fluctuated surface represents that lower stability for each hyperparameter and Monte Carlo trial.

Also, the AUROC surface of the NCAE shows also fluctuated form but it shows less fluctuation in high performance condition than FARED. For summarizing, the stability of NCAE in hyperparameter tuning relatively high to the FARED, and NCAE shows relatively consistent performance in an appropriate hyperparameter such as shown in learning rate at ${1}^{-3}$. 

Moreover, the advantage of NCAE that adjustable hyperparameters diverse than FARED. For example, when constructing a model, NCAE can additionally tune the kernel size as a hyperparameter than FARED when limiting the number of layers, the dimension of the latent space, and the type of optimizer to the same conditions. Thus, the number of NCAE's AUROC surface is more than FARED's, which can be confirmed via comparing Figures~\ref{fig:surfaces}.

\section{Conclusion}
\label{sec:conclusion}
We experimentally demonstrated cutting-edge anomaly detection performance of our NCAE on the road surface via vehicle driving noise. Proposed model, NCAE, beyonds the anomaly detection performance and time efficiency of FARED (4.20\% higher AUROC and 2.99 times faster). We also present the higher and improved stability of NCAE than FARED. Thus, we finally conclude that NCAE offers a cutting-edge architecture for road surface anomaly detection via driving noise. We plan to confirm if NCAE can perform properly to various anomalies such as snow, icy, or some other condition, through additional data collection in the future.

\section*{Acknowledgements}
We are grateful to all the members of our team and to SK Planet Co., Ltd., who have supported this research, not only via data collection, but also by providing equipment for the experiment.

\pagebreak

\vspace{12pt}


\begin{thebibliography}{00}

\bibitem{hall2009guide}
Hall, JW, Smith, Kelly L, Titus-Glover, Leslie, Wambold, James C, Yager, Thomas J, Rado, Zoltan. Guide for pavement friction. Final Report for NCHRP Project 2009; 1: 43.

\bibitem{mcgovern2011state}
McGovern, Colleen M and Rusch, Peter F and Noyce, David A and others. State Practices to Reduce Wet Weather Skidding Crashes. United States. Federal Highway Administration. Office of Safety 2011.

\bibitem{mikolov2010recurrent}
Mikolov, Tom{\'a}{\v{s}} and Karafi{\'a}t, Martin and Burget, Luk{\'a}{\v{s}} and {\v{C}}ernock{\`y}, Jan and Khudanpur, Sanjeev, Recurrent neural network based language model. Eleventh annual conference of the international speech communication association 2010.

\bibitem{graved2012lstm}
Graves, Alex. "Long short-term memory." Supervised sequence labelling with recurrent neural networks. Springer, Berlin, Heidelberg, 2012; 37-45.

\bibitem{park2018fared}
Park, YeongHyeon and Yun, Il Dong. Fast adaptive RNN encoder–decoder for anomaly detection in SMD assembly machine. Sensors 2018; 18(10): 3573.

\bibitem{park2020hpgan}
Park, YeongHyeon, Won Seok Park, and Yeong Beom Kim. Anomaly detection in particulate matter sensor using hypothesis pruning generative adversarial network. ETRI Journal 2020.

\bibitem{schuster1997birnn}
M. Schuster and K. K. Paliwal, Bidirectional recurrent neural networks. IEEE transactions on Signal Processing 1997; 45(11): 2673-2681.

\bibitem{tukey1977exploratory}
Kroese, Dirk P. and Brereton, Tim and Taimre, Thomas and Botev, Zdravko I. Why the Monte Carlo method is so important today. Wiley Interdisciplinary Reviews: Computational Statistics 2014; 6(6): 386-392.

\bibitem{kroese2014monte}
Tukey, John W. Exploratory data analysis. Reading, Mass 1977; 2.
\end{thebibliography}
\end{document}